\documentclass[]{typhoon} 

\usepackage{amsmath,amssymb}
\usepackage[toc,page,header]{appendix}
\usepackage{minitoc}
\usepackage{tikz}
\usetikzlibrary{patterns}
\usepackage{float}
\usepackage{bm}
\usepackage{soul}
\usepackage{algorithm}
\usepackage{algpseudocode}
\usepackage{cleveref}
\usepackage{enumitem}
\usepackage{tabularx}
\setlist[itemize]{leftmargin=*}
\setlist[enumerate]{leftmargin=*}
\usepackage[svgnames]{xcolor}
\definecolor{Gray}{gray}{0.9}
\usepackage{pifont}
\usepackage{xspace}

\newcommand{\huggingface}{\raisebox{-1.5pt}{\includegraphics[height=1.05em]{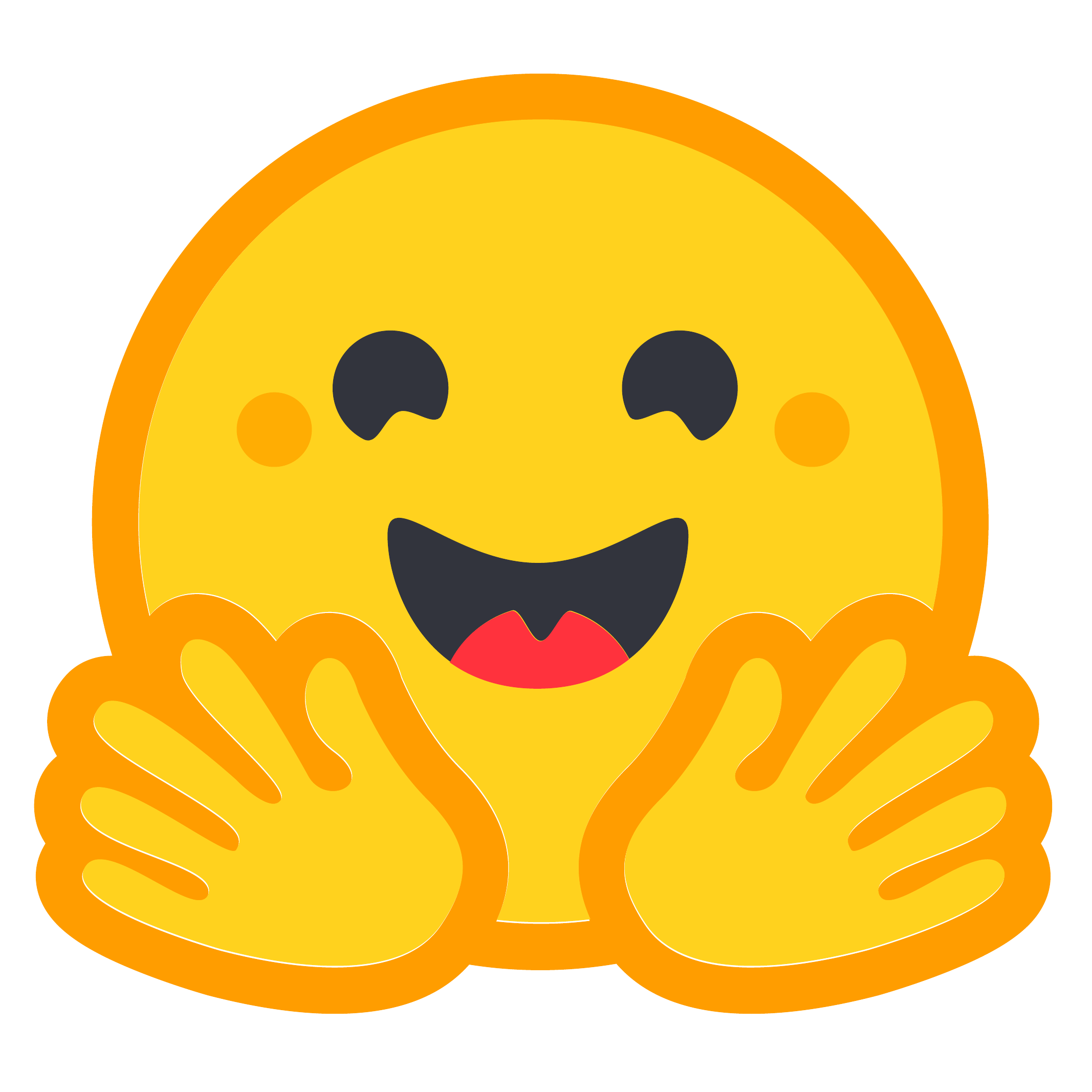}}\xspace}
\newcommand{\github}{\raisebox{-1.5pt}{\includegraphics[height=1.05em]{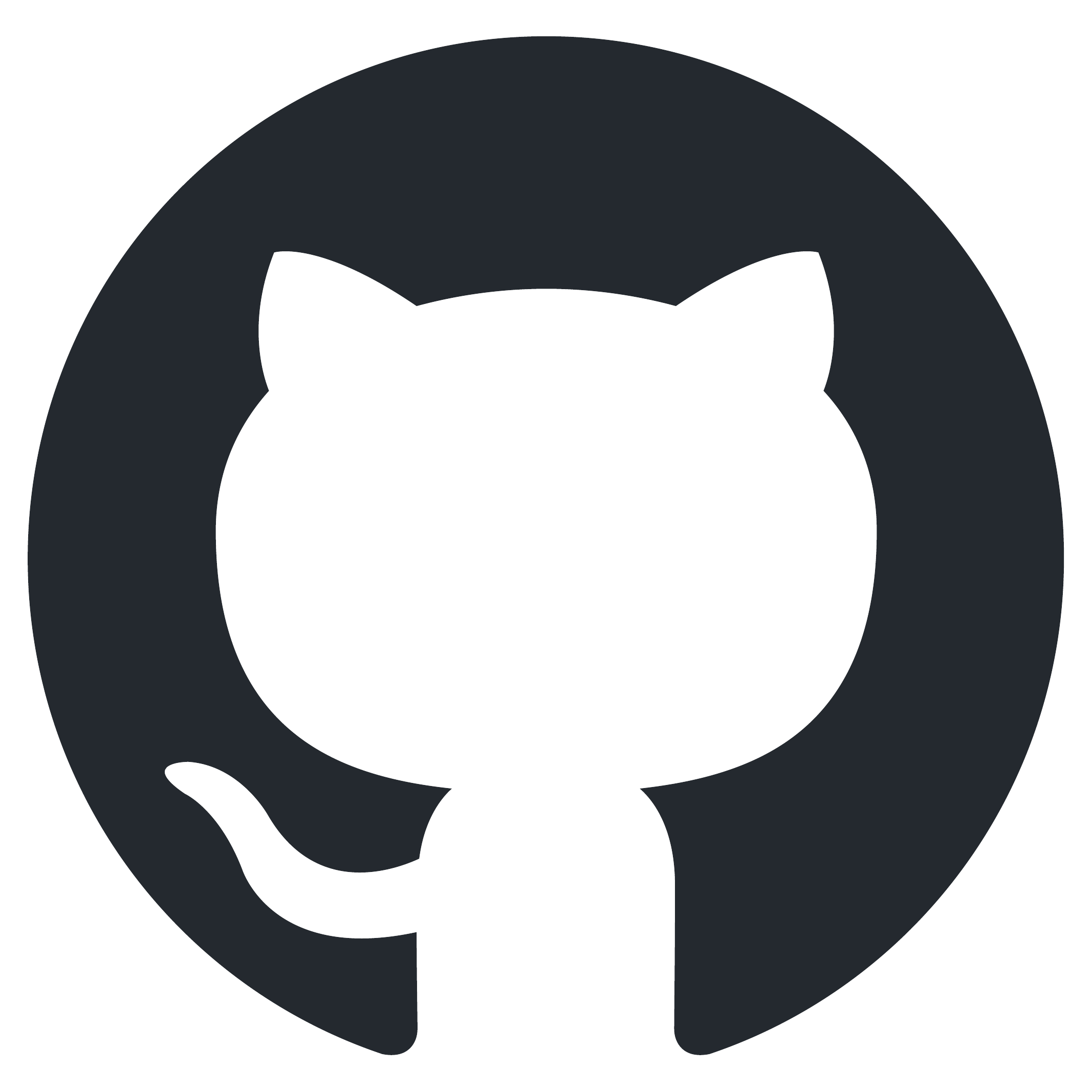}}\xspace}
\newcommand{\typhoon}{\raisebox{-1.5pt}{\includegraphics[height=1.05em]{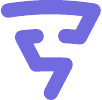}}\xspace}

\title{Typhoon OCR: Open Vision–Language Model For Thai Document Extraction}

\author{
Surapon Nonesung,
Natapong Nitarach,
Teetouch Jaknamon, \\
Pittawat Taveekitworachai, Kunat Pipatanakul
}

\affiliation{
Typhoon, SCB 10X
}

\abstract{
Document extraction is a core component of digital workflows, yet existing vision-language models (VLMs) predominantly favor high-resource languages. Thai presents additional challenges due to script complexity from non-latin letters, the absence of explicit word boundaries, and the prevalence of highly unstructured real-world documents, limiting the effectiveness of current open-source models. This paper presents \textbf{Typhoon OCR}, an open VLM for document extraction tailored for Thai and English. The model is fine-tuned from vision-language backbones using a Thai-focused training dataset. The dataset is developed using a multi-stage data construction pipeline that combines traditional OCR, VLM–based restructuring, and curated synthetic data. Typhoon OCR is a unified framework capable of text transcription, layout reconstruction, and document-level structural consistency. The latest iteration of our model, \textbf{Typhoon OCR V1.5}, is a compact and inference-efficient model designed to reduce reliance on metadata and simplify deployment. Comprehensive evaluations across diverse Thai document categories, including financial reports, government forms, books, infographics, and handwritten documents, show that Typhoon OCR achieves performance comparable to or exceeding larger frontier proprietary models, despite substantially lower computational cost. The results demonstrate that open vision-language OCR models can achieve accurate text extraction and layout reconstruction for Thai documents, reaching performance comparable to proprietary systems while remaining lightweight and deployable.
}

\checkdata[Project Resources]{\newline
\github\texttt{Code:} \url{https://github.com/scb-10x/typhoon-ocr} \newline
\huggingface\texttt{Typhoon OCR 7B}: \url{https://huggingface.co/scb10x/typhoon-ocr-7b} \newline
\huggingface\texttt{Typhoon OCR V1.5 2B}: \url{https://huggingface.co/scb10x/typhoon-ocr1.5-2b} \newline \typhoon\texttt{Blog}: \url{https://opentyphoon.ai/blog/en/typhoon-ocr-release}
}

\begin{document}
\maketitle
\vspace{-20pt}


\section{Introduction}

\begin{figure}[htbp]
    \centering
    \includegraphics[width=0.95\textwidth]{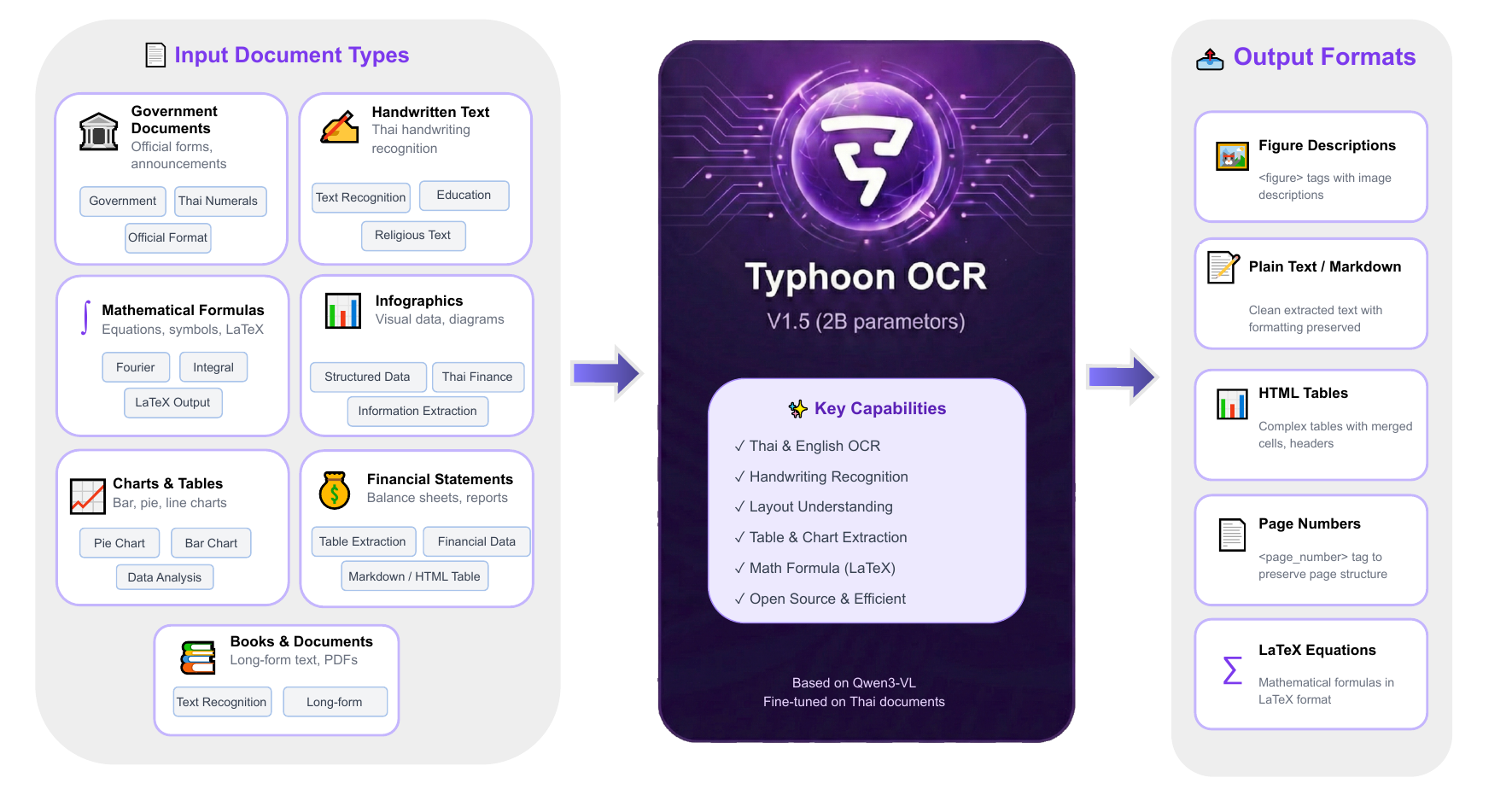}
    \caption{Overview of Typhoon OCR, illustrating supported input document types and the corresponding structured output representations.}
    \label{fig:overview_flow}
\end{figure}

Recent vision-language models (VLMs) are capable of various visual-language tasks from document understanding, integrating optical character recognition, layout analysis, to semantic modeling as a single model \citep{zhang2024visionlanguagemodelsvisiontasks}. While these models demonstrate strong performance on high-resource languages, their effectiveness degrades substantially in low-resource settings, particularly for languages with complex scripts and limited annotated data \citep{nonesung2025thaiocrbenchtaskdiversebenchmarkvisionlanguage}. Thai language is one such case, where linguistic properties and document diversity present persistent challenges for existing models.

The Thai writing system exhibits characteristics that complicate document understanding, including stacked diacritics, context-dependent vowel placement, and the absence of explicit word boundaries, all of which hinder reliable text segmentation and recognition \citep{4600388}. At the document level, Thai materials--such as administrative forms, financial records, receipts, and tabular reports--often contain dense and irregular layouts, further increasing the difficulty of accurate extraction and structural reconstruction.

\Cref{fig:overview_flow} provides an overview of the document types supported by Typhoon OCR and the corresponding output formats. Despite this range of document types and output requirements, general-purpose VLMs, such as Qwen3-VL \citep{wei2025deepseekocrcontextsopticalcompression}, Gemma 3 \citep{gemmateam2025gemma3technicalreport}, and InternVL 3.5 \citep{wang2025internvl35advancingopensourcemultimodal}, which are predominantly trained on high-resource languages, frequently fail to capture these properties, resulting in elevated recognition errors, layout misinterpretation, and semantic inconsistencies.

A primary factor underlying these limitations is Thai data scarcity. The development of large language models (LLMs) for Thai has highlighted the broader challenge of limited high-quality Thai training data; recent Thai LLMs such as the Typhoon series are explicitly designed and evaluated on Thai language datasets to improve Thai language processing \citep{typhoon2}. This issue is more prevalent in multimodal data. In contrast to English or Chinese, Thai lacks large-scale datasets that align document images with structured textual and semantic annotations, limiting the adaptability of existing models.
Consequently, many of existing VLMs have limited exposure to Thai-specific visual and linguistic patterns, restricting their ability to generalize across domains such as public administration, finance, and education. This imbalance is reflected in our evaluations (Tables~\ref{tab:typhoon_eval},~\ref{tab:bleu_by_task}–\ref{tab:lev_by_task}), where proprietary models such as GPT and Gemini achieve competitive performance on some categories but are generally outperformed by Typhoon OCR on structured Thai document extraction tasks.


These challenges motivate the development of a document understanding model explicitly adapted to Thai. Prior research in low-resource optical character recognition (OCR) and multilingual natural language processing suggests that fine-tuning large pretrained models with language- and domain-specific supervision can yield substantial gains without requiring training from scratch. For example, adapting multilingual vision–language transformers to low-resource OCR tasks has been shown to improve recognition performance in non-Latin scripts \citep{cheema2024adapting}, and synthetic data augmentation and fine-tuning have been used to improve OCR performance across diverse low-resource languages \citep{ignat-etal-2022-ocr}. Similarly, transfer learning and fine-tuning of pretrained multilingual models have been effective in low-resource machine translation settings \citep{dalal-etal-2024-ai}, illustrating the broader utility of model adaptation in low-resource linguistic domains.

In this work, we present \textbf{Typhoon OCR}, an open VLM for end-to-end Thai and English document understanding. The model is capable of various visual-language capabilities, including text extraction, layout reconstruction, and document-level semantic modeling. Typhoon OCR is obtained by fine-tuning an open VLM backbone using a task-aligned training corpus constructed from curated real-world documents and synthetic data. By explicitly modeling document structure alongside textual content, our model has capabilities beyond traditional OCR pipelines focused primarily on transcription. Our contributions are as follows:
\begin{itemize}[leftmargin=*]
\setlength\itemsep{-0.1em}
    \item We introduce Typhoon OCR, an open VLM tailored to Thai document extraction.
    \item We propose a data curation and synthesis pipeline that mitigates multimodal data scarcity for Thai.
    \item We comprehensively evaluate our models against the baselines, demonstrating improvements over existing open baselines and competitive performance relative to proprietary models on Thai document benchmarks.
    \item Our models and associated resources are released under permissive licenses to support reproducible research and practical deployment.
\end{itemize}

\section{Typhoon OCR}
Typhoon OCR is an open VLM for end-to-end document extraction in Thai and English, including text recognition and layout-aware content reconstruction. It targets real-world documents with complex and heterogeneous layouts, where existing OCR pipelines and general-purpose VLMs perform poorly in low-resource settings.

The model is trained by fine-tuning a vision-language backbone on a task-aligned corpus combining curated real documents and synthetic data. Experiments show that Typhoon OCR substantially outperforms existing open-source methods and achieves competitive performance relative to proprietary models on Thai document benchmarks.

\subsection{Data}
\subsubsection{Dataset Creation Pipeline}
\begin{table}[h]
\centering
\begin{tabular}{@{}p{4cm}p{5.5cm}p{5.5cm}@{}}
\toprule
\textbf{Feature} & \textbf{Default Mode} & \textbf{Structure Mode} \\
\midrule
\textbf{Target Documents} &
Loosely structured or unstructured documents, such as receipts, menus, tickets, infographics, and handwritten notes. &
Highly structured documents with complex layouts, including financial reports, academic papers, government forms, and books. \\
\midrule
\textbf{Layout Modeling} &
Lightweight layout preservation with an emphasis on content continuity and readability. &
Explicit structural parsing with detailed reconstruction of hierarchical and semantic regions. \\
\midrule
\textbf{Output Representation} &
Only \texttt{Markdown} with minimal layout annotations. &
\texttt{Markdown} for narrative text, \texttt{HTML} for complex tables and structured layouts, and \texttt{<figure>} tags for visual elements. \\
\bottomrule
\end{tabular}
\caption{Comparison of supervision modes used in Typhoon OCR.}
\label{tab:modes-comparison}
\end{table}

To support different types of documents, we construct a training corpus that allows the model to operate in two modes: \emph{Default Mode} and \emph{Structure Mode}. These modes differ in how much layout information is preserved in the output. A single supervision format is not suitable for both loosely structured documents (such as receipts or handwritten notes) and highly structured documents (such as financial reports or government forms). Using one format for all cases would either add unnecessary complexity to simple documents or fail to capture important layout details in complex ones.

We therefore adopt two modes as a simple and practical division based on document structure: documents with little or weak layout structure, and documents with clear and complex layout organization. This separation covers most document types in our dataset without introducing excessive design complexity.

\begin{figure}[htbp]
    \centering
    \includegraphics[width=\linewidth]{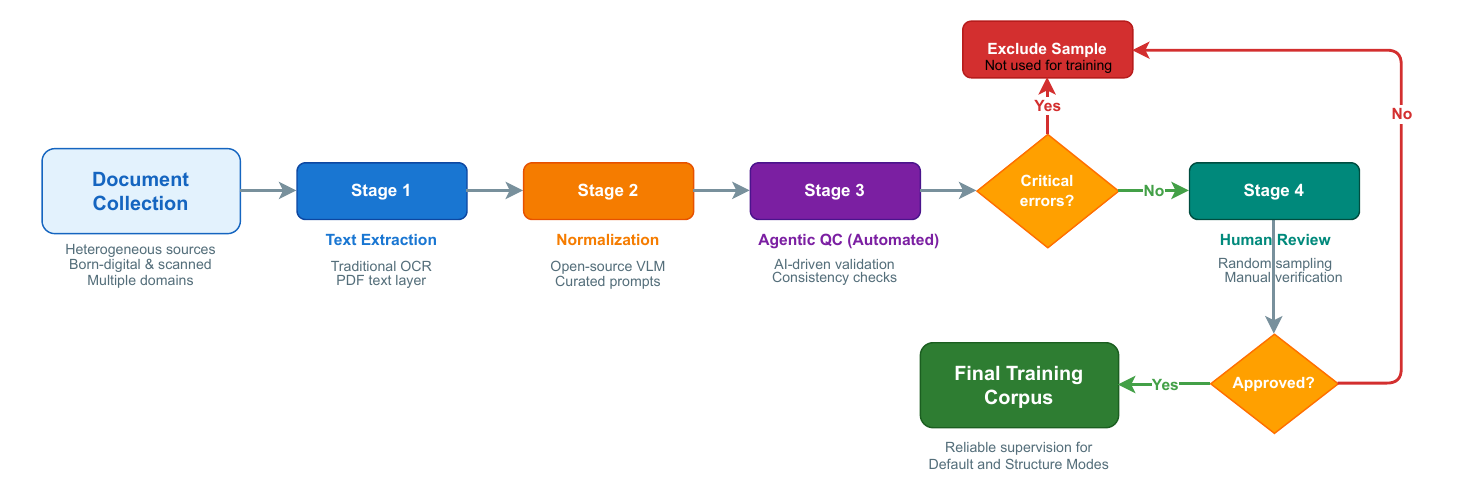}
    \caption{Overview of the multi-stage dataset construction pipeline used to generate training data for Typhoon OCR under Structure Mode.}
    \label{fig:dataset_flow}
\end{figure}

This design also follows the availability of training data. Default Mode reuses general document annotations from the Typhoon2 Vision training corpus~\citep{typhoon2}. Structure Mode, in contrast, is newly created using the dataset construction pipeline shown in \Cref{fig:dataset_flow}, which produces structure-aware annotations for complex documents. In practice, users choose the mode based on the document type, as summarized in \Cref{tab:modes-comparison}.


As shown in \Cref{fig:dataset_flow}, the training data are collected from diverse sources, including digital-native documents and scanned materials spanning multiple domains. Rather than applying a single processing strategy, we use a multi-stage pipeline that incrementally refines annotations while balancing scalability and supervision quality. These stages are:

\paragraph{Stage 1} Textual content is extracted using conventional OCR systems and, where available, document text-layer parsing. This stage focuses on reliable character- and word-level transcription, particularly for clean digital documents and high-resolution scans, and provides a stable low-level supervision signal.

\paragraph{Stage 2} The extracted text is reorganized using open-source VLMs guided by structured prompts. This step transforms raw OCR outputs into coherent document representations by incorporating layout-aware formatting, section grouping, and basic semantic organization.

\paragraph{Stage 3} Automated quality control is applied through agent-based consistency checks. These checks detect common failure modes, including structural inconsistencies, missing or duplicated content, incorrect ordering, and misalignment between visual input and generated annotations.

\paragraph{Stage 4} A subset of samples is selected for human verification. Annotators assess annotation fidelity with respect to the source documents, and samples exhibiting substantial or irrecoverable errors are removed from the corpus.

This multi-stage pipeline enables scalable dataset construction while mitigating noise from automated annotation. As a result, the final corpus provides consistent, high-quality supervision for fine-tuning Typhoon OCR under both modes.

\subsubsection{Dataset Statistics}

\begin{figure}[htbp]
    \centering
    \begin{subfigure}{0.45\textwidth}
        \centering
        \includegraphics[width=\textwidth]{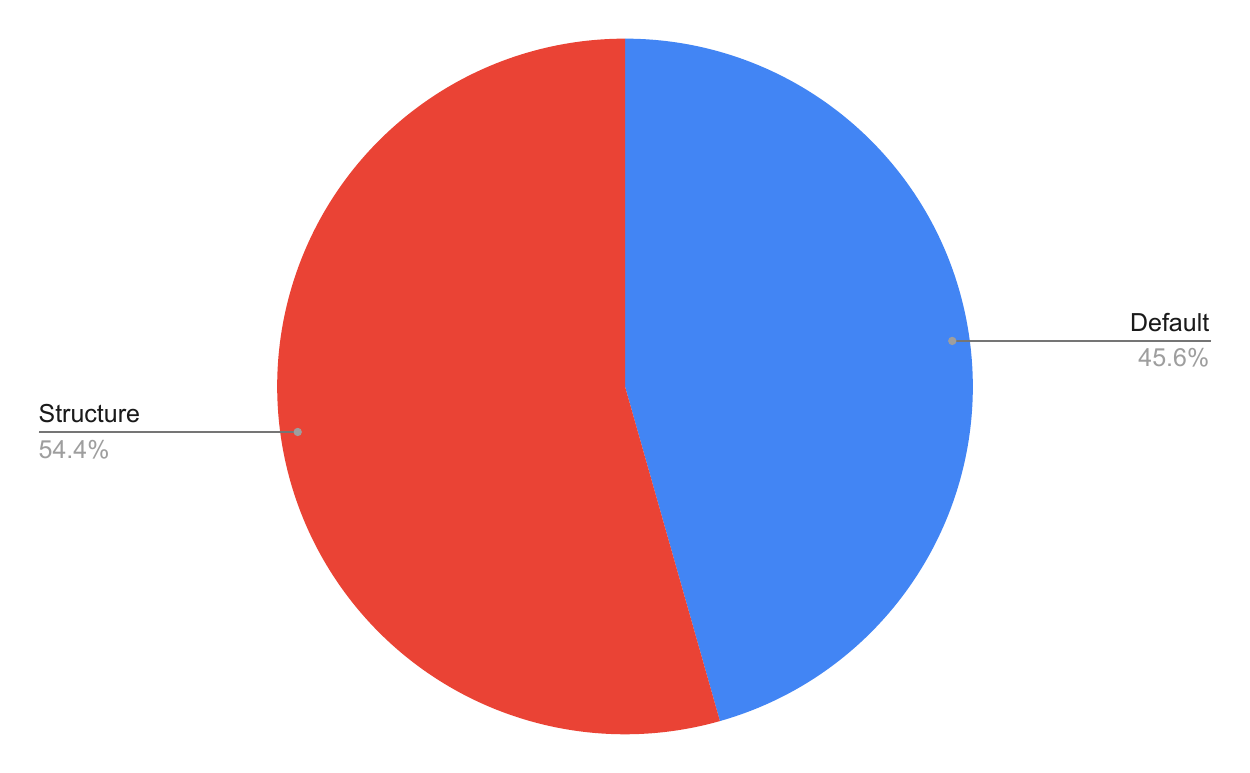}
        \caption{Distribution of training samples across modes}
        \label{fig:dataset_dis}
    \end{subfigure}
    \hfill
    \begin{subfigure}{0.5\textwidth}
        \centering
        \includegraphics[width=\textwidth]{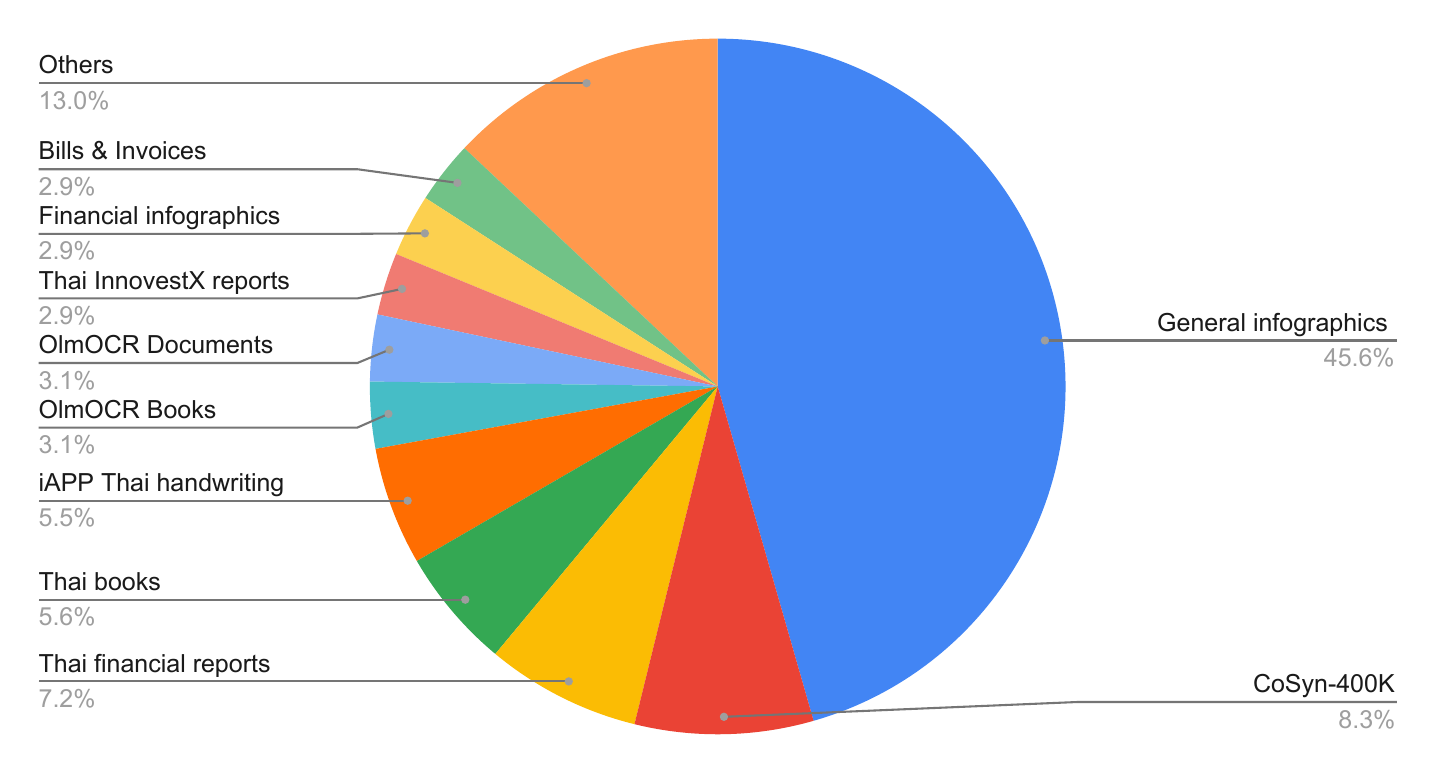}
        \caption{Distribution of training samples by data source}
        \label{fig:dataset_by}
    \end{subfigure}
    \caption{Composition of the training corpus used for Typhoon OCR. Figure~\ref{fig:dataset_dis} shows the allocation of samples between Default Mode and Structure Mode supervision, while Figure~\ref{fig:dataset_by} presents the relative contributions of different data sources.}
\end{figure}

\Cref{fig:dataset_dis} summarizes the composition of the training corpus used to fine-tune Typhoon OCR. The dataset is curated from publicly available sources in both Thai and English and spans a wide range of document types, including infographics, government forms, financial reports, books, and handwritten materials. This diversity is intended to reflect realistic variation in document layout, visual quality, and linguistic content encountered in practical document understanding tasks.

As shown in \Cref{fig:dataset_by}, the corpus consists of a heterogeneous mixture of document sources across both languages. General infographic-style documents constitute the largest portion of the dataset (45.6\%), providing broad coverage of diverse layouts and visual styles. Structured synthetic documents from the CoSyn-400K dataset\footnote{\url{https://huggingface.co/datasets/allenai/CoSyn-400K}} \citep{yang2025scaling} account for 8.3\% of the corpus and provide controlled layout variations that facilitate learning of document structure.

Domain-specific Thai documents form a substantial component of the dataset. These include financial reports published by Thai SET organizations (7.2\%), digitized Thai books spanning multiple genres and formatting conventions (5.6\%), and handwritten materials sampled from the iAPP Thai handwriting dataset\footnote{\url{https://huggingface.co/datasets/iapp/thai_handwriting_dataset}} (5.5\%). For the handwriting data, subsampling is applied to mitigate the impact of known labeling inconsistencies.

Additional scanned documents and book pages are drawn from the olmOCR-mix-0225 collection\footnote{\url{https://huggingface.co/datasets/allenai/olmOCR-mix-0225}}, contributing a combined 6.2\% of the corpus. The remaining portion of the dataset comprises Thai InnovestX reports, financial infographics, and bills and invoices (8.7\%), alongside a long-tail category (13.0\%) that includes government forms, certificates, academic papers, and other miscellaneous scanned documents. In total, the training corpus contains 77,029 document samples.

\subsection{Experimental Setup}

\subsubsection{Training}

Typhoon OCR is trained using full-parameter supervised fine-tuning (SFT) on the Qwen2.5-VL model family \citep{bai2025qwen25vltechnicalreport}, covering both the 3B and 7B parameter variants. The training pipeline is based on the open-source \texttt{olmOCR} framework\footnote{\url{https://github.com/allenai/olmocr}} \citep{poznanski2025olmocrunlockingtrillionstokens}, with extensions to support multilingual document understanding and long-context modeling. The models are trained on 4$\times$H100 GPUs for three epochs, and the final checkpoint is selected based on performance on a held-out validation set.

Input document images are resized to a fixed width of 1,800 pixels, providing a trade-off between visual fidelity and computational efficiency. The anchor text length is set to 8,000 tokens to balance layout coverage and memory constraints, and the maximum sequence length is limited to 17,000 tokens, allowing the model to process long-form documents and dense layouts without truncation.


\subsubsection{Evaluation}\label{sec:v1_eval}

\paragraph{Metrics}
We evaluate Typhoon OCR using standard metrics from OCR and natural language generation (NLG) that capture complementary aspects of output quality. Lexical accuracy is assessed using \textbf{BLEU} \citep{papineni-etal-2002-bleu}, which measures $n$-gram overlap between predicted and reference texts. Structural and sequential similarity is measured using \textbf{ROUGE-L} \citep{lin-2004-rouge}, based on the longest common subsequence. Although originally developed for NLG, these metrics correlate well with document extraction in structured outputs. Character-level transcription fidelity is quantified using \textbf{Levenshtein distance} \citep{haldar2011levenshteindistancetechniquedictionary}, where lower values indicate fewer edit operations.

\paragraph{Benchmark}
Evaluation is conducted on an in-house Thai document corpus designed to reflect real-world document variability in both layout and content. The corpus comprises three categories: Thai financial reports, which include complex tables, charts, and multilingual content; Thai government forms, characterized by dense layouts, domain-specific terminology, and handwritten annotations; and Thai books, consisting of long-form text interleaved with figures, diagrams, and other visual elements.

\paragraph{Protocol}

Evaluation is conducted under two input conditions to examine robustness across document representations. In the \emph{PDF with metadata} setting, the model is provided with native PDF information, including text layers and layout annotations, when available. In the \emph{image-only} setting, the model receives rasterized document images without access to structural metadata.

This setup isolates the contribution of explicit layout information and allows analysis of the model's ability to infer document structure directly from visual input.

\begin{table}[htbp]
    \centering
    \footnotesize
    \renewcommand*{\arraystretch}{1.2}
    \begin{tabular}{lccc}
        \toprule
        \textbf{Model} & \textbf{BLEU}$\uparrow$ & \textbf{ROUGE-L}$\uparrow$ & \textbf{Levenshtein}$\downarrow$ \\
        \midrule
        \rowcolor{typhoonpurple!20}
        \multicolumn{4}{l}{\textbf{Thai Financial Reports}} \\
        GPT-4o (2024-11-20) & 0.25 & 0.51 & 0.56 \\
        Gemini 2.5 Flash (2025-04-17) & 0.52 & 0.70 & 0.35 \\
        Typhoon OCR 3B (PDF) & 0.90 & 0.93 & 0.08 \\
        Typhoon OCR 3B (Image) & 0.90 & 0.93 & \textbf{0.07} \\
        Typhoon OCR 7B (PDF) & \textbf{0.91} & \textbf{0.94} & \textbf{0.07} \\
        Typhoon OCR 7B (Image) & \textbf{0.91} & \textbf{0.94} & 0.08 \\
        \midrule
        \rowcolor{typhoonpurple!20}
        \multicolumn{4}{l}{\textbf{Thai Government Forms}} \\
        GPT-4o (2024-11-20) & 0.25 & 0.45 & 0.57 \\
        Gemini 2.5 Flash (2025-04-17) & 0.74 & 0.87 & 0.15 \\ 
        Typhoon OCR 3B (PDF) & 0.92 & \textbf{0.96} & 0.05 \\
        Typhoon OCR 3B (Image) & \textbf{0.93} & \textbf{0.96} & \textbf{0.04} \\
        Typhoon OCR 7B (PDF) & 0.89 & 0.94 & 0.08 \\
        Typhoon OCR 7B (Image) & 0.89 & 0.94 & 0.07 \\
        \midrule
        \rowcolor{typhoonpurple!20}
        \multicolumn{4}{l}{\textbf{Thai Books}} \\
        GPT-4o (2024-11-20) & 0.34 & 0.49 & 0.59 \\
        Gemini 2.5 Flash (2025-04-17) & 0.47 & 0.59 & 0.47 \\
        Typhoon OCR 3B (PDF) & 0.63 & 0.71 & 0.32 \\
        Typhoon OCR 3B (Image) & \textbf{0.64} & 0.71 & 0.32 \\
        Typhoon OCR 7B (PDF) & 0.63 & 0.71 & \textbf{0.31} \\
        Typhoon OCR 7B (Image) & \textbf{0.64} & \textbf{0.72} & \textbf{0.31} \\
        \bottomrule
    \end{tabular}
    \caption{Performance comparison on Thai document parsing in Structure Mode. Higher BLEU and ROUGE-L and lower Levenshtein indicate better performance.}
    \label{tab:typhoon_eval}
\end{table}

\subsection{Results and Discussion}

\Cref{tab:typhoon_eval} reports performance of Typhoon OCR (3B and 7B) relative to GPT-4o and Gemini 2.5 Flash across document categories. Typhoon OCR consistently outperforms baseline models on financial reports and government forms, which exhibit dense layouts and structured content. These gains are most pronounced when PDF metadata are available, indicating that explicit layout cues contribute to improved structural reconstruction.

Performance on the Thai Books subset is lower across all models. This category introduces additional complexity due to frequent visual elements, such as illustrations and non-standard figures, which increase ambiguity in figure representation and layout interpretation. The results suggest that figure understanding remains a limitation in current VLMs.

The performance difference between PDF-based and image-only inputs is small for Typhoon OCR, indicating effective alignment between visual and textual representations. Notably, the 3B variant achieves results comparable to the 7B model on several tasks, particularly government forms, suggesting that smaller models can be effective under constrained deployment settings.

During our pilot experiments, we observe that training with \emph{heterogeneous} image resolutions leads to less stable optimization and reduced accuracy. Standardizing inputs by resizing images to a fixed width of 1,800 pixels while preserving aspect ratio, as we applied to Typhoon OCR, improves both training stability and evaluation performance. This finding aligns with prior observations that OCR and document understanding are sensitive to resolution variability, especially in low-resource training regimes \citep{xiao2025scalinglanguagecentricomnimodalrepresentation}. 



\section{Typhoon OCR V1.5}

While Typhoon OCR has received a lot of positive feedback, post-deployment analysis of Typhoon OCR revealed some limitations that affect its usability and efficiency in real-world settings. First, dependence on PDF anchor metadata introduced increased inference latency, particularly for long documents and those with complex layouts. Second, the separation of operating modes increased user-facing complexity and led to performance variability when modes were not appropriately selected. Third, although the 3B and 7B models are relatively small compared to frontier VLMs, further reducing their computational footprint would significantly increase their applicability in latency-sensitive and resource-constrained environments. Finally, although the training corpus covered diverse document types, further expansion in data diversity and output representations was necessary to improve robustness and generalization.

Typhoon OCR V1.5 addresses these limitations through data and training refinements aimed at improving inference efficiency, simplifying the interaction interface, and strengthening robustness across document types and deployment scenarios.

\subsection{Data}
\subsubsection{Dataset Creation Pipeline}

In Typhoon OCR V1, dataset construction relied on separate mode strategies for different operating modes and on PDF anchor metadata for annotation. In V1.5, this design is simplified by adopting a single unified mode, eliminating the need for mode selection during training and inference and enabling annotations to be generated directly from visual input.

Annotation quality is further improved by using stronger and more recent multilingual labeling models--specifically Qwen3-VL~\citep{bai2025qwen3vltechnicalreport} and Dots.OCR~\citep{li2025dotsocrmultilingualdocumentlayout}--which provide enhanced multilingual and Thai document understanding capabilities. In addition, the training corpus is expanded to cover a wider range of document types and layouts.

Beyond real-world documents, we incorporate two additional data sources to address complementary limitations. First, Thai-translated visual question answering (VQA) data is included to preserve general vision--language grounding and basic multimodal reasoning ability in Thai, preventing over-specialization to document transcription alone. Second, synthetic documents are generated to compensate for the scarcity of annotated Thai documents containing complex layouts, mathematical expressions, and charts, and to increase coverage of rare vocabulary and typographic variations.

Overall, while the data construction pipeline follows that of the original Typhoon OCR (Figure~\ref{fig:dataset_flow}), these changes reduce pipeline complexity and improve robustness and consistency across diverse document extraction scenarios. Next, we detail the data processing and synthesis pipelines for the additional data sources used in Typhoon OCR V1.5.

\paragraph{Thai VQA}
\textbf{The Cauldron}\footnote{\url{https://huggingface.co/datasets/HuggingFaceM4/the_cauldron}} \citep{laurençon2024matters}, containing diverse visual question answering tasks covering object recognition, spatial reasoning, and basic document understanding, but is primarily available in English. To improve the model's performance in Thai, we randomly sample VQA instances and translate both questions and answers into Thai using \textbf{Typhoon Translate}\footnote{\url{https://huggingface.co/scb10x/typhoon-translate-4b}}.

\paragraph{Synthetic Data}

Synthetic documents are introduced to address gaps in real-world Thai document coverage, particularly for mathematical notation, rare vocabulary, typographic variation, and complex visual elements. As shown in \Cref{fig:dataset_synth_ocr}, we construct a document generation pipeline with the following stages:

\begin{figure}[htbp]
    \centering
    \includegraphics[width=\linewidth]{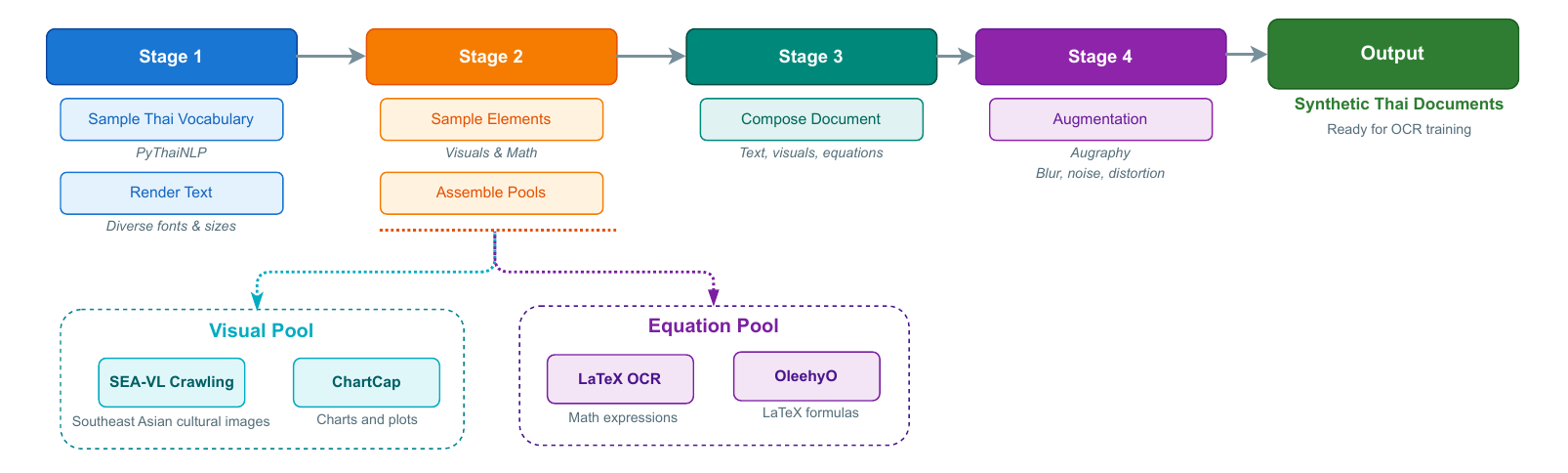}
    \caption{Multi-stage pipeline for generating synthetic Thai document images for OCR training.}
    \label{fig:dataset_synth_ocr}
\end{figure}

\subparagraph{Stage 1} Thai vocabulary is randomly sampled from PyThaiNLP \citep{phatthiyaphaibun-etal-2023-pythainlp} and rendered using diverse font families and font sizes to increase robustness to typographic variation and low-frequency lexical forms

\subparagraph{Stage 2} Visual elements are sampled from SEA-VL Crawling\footnote{\url{https://huggingface.co/datasets/SEACrowd/sea-vl_crawling}} \citep{cahyawijaya2025crowdsourcecrawlgeneratecreating}, which contains culturally relevant images from Southeast Asia, and from ChartCap\footnote{\url{https://huggingface.co/datasets/junyoung-00/ChartCap}} \citep{lim2025chartcapmitigatinghallucinationdense} to provide real-world charts and plots for training layout and figure understanding.

\subparagraph{Stage 3} Mathematical expressions are sampled from LaTeX OCR\footnote{\url{https://huggingface.co/datasets/linxy/LaTeX_OCR}} and OleehyO LaTeX Formulas\footnote{\url{https://huggingface.co/datasets/lamm-mit/OleehyO-latex-formulas}} to improve recognition of equation structures and symbolic layouts.

\subparagraph{Stage 4} We apply document-level image augmentation using Augraphy\footnote{\url{https://github.com/sparkfish/augraphy}} \citep{augraphy_paper} to simulate real-world acquisition artifacts such as blur, noise, compression, illumination variation, and geometric distortion. This step increases visual diversity and improves robustness to degraded scanning and photographic conditions.

These components are combined to generate documents containing mixed textual, mathematical, and visual content with controlled layout variation and Thai-specific linguistic features.

\subsubsection{Dataset Statistics}

\begin{figure}[htbp]
    \centering
    \includegraphics[width=0.6\textwidth]{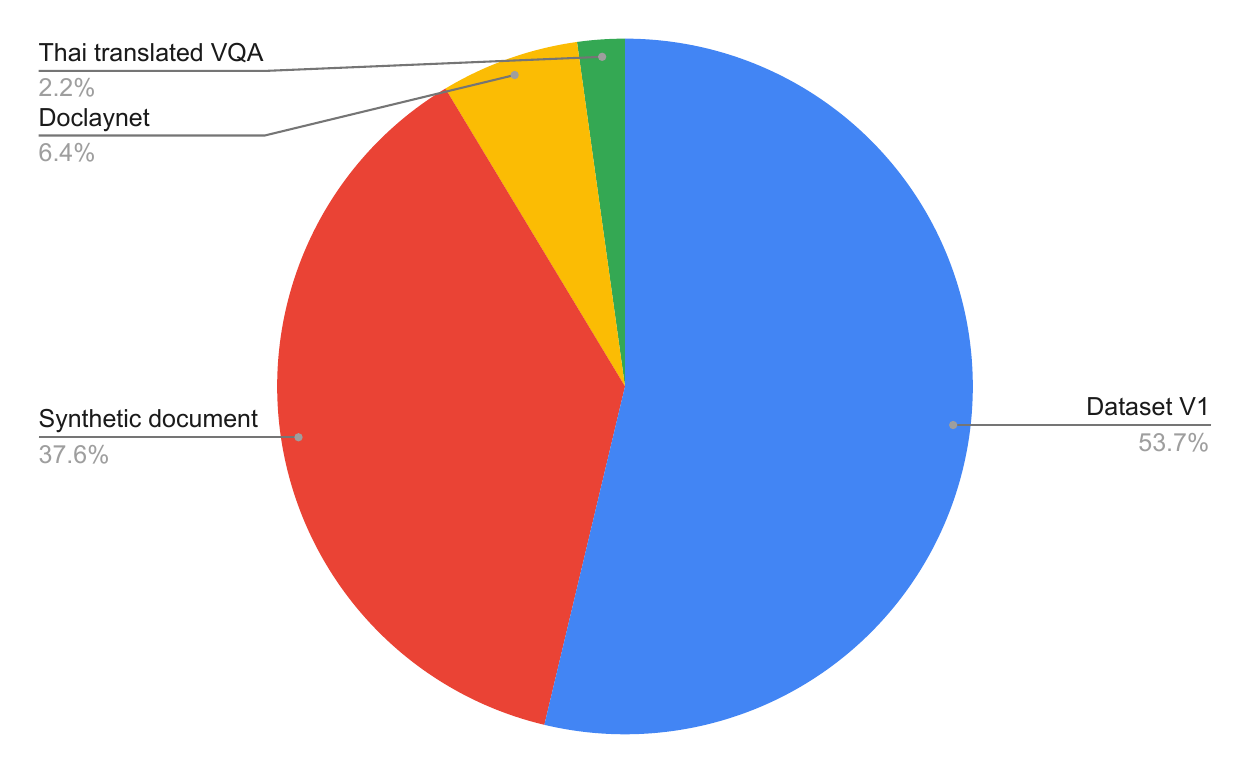}
    \caption{Training dataset distribution for Typhoon OCR V1.5}
    \label{fig:dataset_dis2}
\end{figure}

\Cref{fig:dataset_dis2} summarizes the statistics of training corpus used to fine-tune Typhoon OCR V1.5. The training corpus consists of a mixture of Thai and English document sources.

The largest portion of the dataset (53.7\%) is retained from the \textbf{Typhoon OCR V1 training corpus} to ensure performance consistency across model versions and preserve coverage of previously observed document distributions. The dataset also include VQA samples from The Cauldron, constituting 2.2\% of the overall training corpus and is intentionally kept as a small fraction, as Typhoon OCR is primarily designed for document understanding rather than general-purpose VQA. The inclusion of this subset aims to preserve basic multimodal reasoning capabilities and mitigate catastrophic forgetting of generic vision-language knowledge during document-centric fine-tuning.

Structured document supervision is supplemented using a subset of DocLayNet-v1.2\footnote{\url{https://huggingface.co/datasets/docling-project/DocLayNet-v1.2}} \citep{doclaynet2022}, contributing 6.4\% of the corpus and providing high-quality annotations for layout segmentation and structural regions. The remaining 37.6\% consists of synthetic documents generated using the pipeline described in the previous section. We intentionally maintain a substantial synthetic portion to compensate for the limited availability of annotated Thai documents containing equations and charts. In total, the
training corpus contains 155,403 document samples.

\subsection{Experimental Setup}
\subsubsection{Training}

Typhoon OCR V1.5 is trained using full-parameter SFT on the Qwen3-VL 2B VLM. The training framework is based on the open-source \texttt{Axolotl} framework\footnote{\url{https://github.com/axolotl-ai-cloud/axolotl}} \citep{axolotl}, with extensions to support long-context multimodal inputs and document-centric training objectives.

During preprocessing, a resolution-aware strategy is applied to balance visual fidelity and computational efficiency. Images with a maximum dimension below 1,800 pixels are \emph{retained} at their original resolution, while larger images are resized to a maximum width of 1,800 pixels with aspect ratio preserved. The maximum sequence length is set to 16,384 tokens to accommodate long documents and complex layouts.

Training is performed on 4$\times$H100 GPUs for two epochs, and the final checkpoint is selected based on validation performance. Quantization-aware training \citep{Jacob_2018_CVPR} is applied during fine-tuning to expose the model to quantization effects, enabling efficient low-precision inference with only limited impact on accuracy.

\subsubsection{Evaluation}

\paragraph{Protocol and Metrics} We evaluate Typhoon OCR V1.5 using the same metrics and protocols as V1 (\Cref{sec:v1_eval}) to ensure comparability across model iterations. Namely, we report \textbf{BLEU}, \textbf{ROUGE-L}, and \textbf{Levenshtein distance} to measure lexical accuracy, sequence-level structural similarity, and character-level transcription fidelity, respectively.

\paragraph{Benchmark}

Evaluation is conducted on held-out test sets from Typhoon OCR V1, which are further refined through additional human annotation and quality control to reduce noise and improve consistency. In addition, two new evaluation sets are introduced to assess performance on document types that were underrepresented in V1.

The evaluation corpus comprises six categories:
\begin{enumerate}
    \item \textbf{Thai Books:} Consist of long-form documents with extended text interleaved with figures and illustrations.
    \item \textbf{Thai Government Forms:} Reused from V1 with enhanced human verification. This benchmark includes dense layouts, tables, charts, and handwritten annotations.
    \item \textbf{Thai Financial Reports:} Also reused from V1 with enhanced human verification.
    \item \textbf{Infographics:} Contain loosely structured documents with prominent visual elements.
    \item \textbf{Handwritten Forms:} Consist of administrative and financial forms that combine handwritten and printed text.
    \item \textbf{Others:} A diverse collection of semi-structured and unstructured documents, including receipts, bills, tickets, and miscellaneous transactional records.
\end{enumerate}
\begin{table}[htbp]
\centering
\renewcommand*{\arraystretch}{1.2}
\resizebox{\linewidth}{!}{
\begin{tabular}{lcccc}
\toprule
\textbf{Task} & \textbf{Gemini 2.5 Pro} & \textbf{GPT-5} & \textbf{Typhoon OCR V1 7B} & \textbf{Typhoon OCR V1.5 2B} \\
\midrule
Thai Books             & 0.512 & 0.710 & 0.708 & \textbf{0.746} \\
Thai Government Forms  & 0.797 & 0.569 & 0.849 & \textbf{0.870} \\
Thai Financial Reports & 0.657 & 0.457 & \textbf{0.849} & 0.819 \\
Infographics           & \textbf{0.465} & 0.297 & 0.246 & 0.408 \\
Handwriting Forms      & \textbf{0.594} & 0.368 & 0.321 & 0.522 \\
Others                 & \textbf{0.603} & 0.352 & 0.376 & 0.499 \\
\midrule
\textbf{Average}       & 0.605 & 0.459 & 0.558 & \textbf{0.644} \\
\bottomrule
\end{tabular}
}
\caption{BLEU scores by document category (higher is better).}
\label{tab:bleu_by_task}
\end{table}
\begin{table}[htbp]
\centering
\renewcommand*{\arraystretch}{1.2}
\resizebox{\linewidth}{!}{
\begin{tabular}{lcccc}
\toprule
\textbf{Task} & \textbf{Gemini 2.5 Pro} & \textbf{GPT-5} & \textbf{Typhoon OCR V1 7B} & \textbf{Typhoon OCR V1.5 2B} \\
\midrule
Thai Books             & 0.676 & 0.922 & 0.871 & \textbf{0.949} \\
Thai Government Forms  & 0.894 & 0.706 & 0.942 & \textbf{0.967} \\
Thai Financial Reports & 0.757 & 0.603 & \textbf{0.933} & 0.910 \\
Infographics           & \textbf{0.677} & 0.481 & 0.373 & 0.527 \\
Handwriting Forms      & \textbf{0.739} & 0.514 & 0.454 & 0.645 \\
Others                 & \textbf{0.716} & 0.482 & 0.541 & 0.645 \\
\midrule
\textbf{Average}       & 0.743 & 0.618 & 0.686 & \textbf{0.774} \\
\bottomrule
\end{tabular}
}
\caption{ROUGE-L scores by document category (higher is better).}
\label{tab:rouge_by_task}
\end{table}

\begin{table}[htbp]
\centering
\renewcommand*{\arraystretch}{1.2}
\resizebox{\linewidth}{!}{
\begin{tabular}{lcccc}
\toprule
\textbf{Task} & \textbf{Gemini 2.5 Pro} & \textbf{GPT-5} & \textbf{Typhoon OCR V1 7B} & \textbf{Typhoon OCR V1.5 2B} \\
\midrule
Thai Books             & 0.334 & 0.084 & 0.136 & \textbf{0.053} \\
Thai Government Forms  & 0.096 & 0.267 & 0.065 & \textbf{0.035} \\
Thai Financial Reports & 0.256 & 0.356 & 0.082 & \textbf{0.079} \\
Infographics           & \textbf{0.380} & 0.561 & 0.671 & 0.544 \\
Handwriting Forms      & \textbf{0.327} & 0.533 & 0.556 & 0.416 \\
Others                 & \textbf{0.342} & 0.540 & 0.480 & 0.377 \\
\midrule
\textbf{Average}       & 0.289 & 0.390 & 0.332 & \textbf{0.251} \\
\bottomrule
\end{tabular}
}
\caption{Levenshtein distance by document category (lower is better).}
\label{tab:lev_by_task}
\end{table}

\subsection{Results and Discussion}

\Cref{tab:bleu_by_task,tab:rouge_by_task,tab:lev_by_task} report a comparative evaluation of Typhoon OCR V1.5 against Typhoon OCR V1 and two frontier VLM baselines across multiple Thai document categories. We analyze performance trends in relation to model design and document characteristics. Across all metrics, Typhoon OCR V1.5 achieves higher average performance than Typhoon OCR V1, despite having fewer parameters (2B versus 7B). This result indicates that improved data and training recipe have a greater impact on document extraction quality than model size alone. From a deployment perspective, the results suggest that compact, task-adapted models can provide strong performance while reducing computational overhead compared to \textit{one-size-fits-all} proprietary models.

Performance gains are most pronounced for structured document types, including Thai government forms and financial reports. In these categories, Typhoon OCR V1.5 consistently outperforms proprietary baselines on BLEU and ROUGE-L and achieves lower Levenshtein distances. This behavior reflects the benefit of explicit layout modeling and domain-aligned supervision for documents with regular structural patterns, which are common in administrative and financial workflows.

For visually heterogeneous categories such as infographics and handwritten forms, proprietary models achieve lower character-level error rates. Nevertheless, Typhoon OCR V1.5 substantially improves over V1, narrowing the gap in both lexical and structural metrics. This is an area for improvement that we aim to improve in future iterations.

\section{Conclusion}

This report introduces \textbf{Typhoon OCR}, a family of VLMs designed to address limitations in document understanding for Thai. The models consistently show improvements over the base model across various document understanding tasks, including transcription accuracy, layout reconstruction, and structural consistency, and perform competitively with frontier proprietary systems. Notably, Typhoon OCR V1.5 achieves these gains with a smaller model of just 2B parameters, reducing inference cost while matching or exceeding the performance of larger proprietary models across multiple document categories. These results demonstrate that robust document understanding in low-resource settings can be achieved through targeted adaptation of pretrained VLMs and carefully designed data pipelines, without training models from scratch or relying on closed systems, making this approach suitable for resource-constrained and privacy-sensitive deployments.

Several limitations remain. Performance degrades on severely degraded inputs, such as low-resolution images, motion blur, and occlusions, suggesting the need for improved data recipes or explicit modeling of noise and capture artifacts. In addition, although the models currently support primarily Thai and English, extending them to other low-resource languages is a natural direction for future research. Moreover, the current Typhoon OCR model series focuses on document extraction and does not explicitly support higher-level reasoning tasks. Future work will extend the framework to applications such as diagram understanding and structured information extraction. While our evaluation is suitable for the current development stage, broader assessment on academic benchmarks, such as ThaiOCRBench~\citep{nonesung2025thaiocrbenchtaskdiversebenchmarkvisionlanguage}, is a natural next step for better understanding model capabilities.

\section*{Acknowledgments}
Beyond the primary authors, we gratefully acknowledge the Typhoon Team members at SCB 10X whose contributions made this project possible: Adisai Na-Thalang, Chanakan Wittayasakpan, Kritsadha Phatcharoen, Tanawin Samutsin, Shah Faisal Wani, Sittipong Sripaisarnmongkol, Warit Sirichotedumrong, Potsawee Manakul, Krisanapong Jirayoot, Oravee Smithiphol, Kasima Tharnpipitchai, and Kaweewut Temphuwapat. We also extend our appreciation to the SCBx R\&D Team for their support, resources, and valuable insights. Lastly, we are grateful to the global and local AI communities for open-sourcing resources and sharing knowledge.

\bibliographystyle{plainnat}
\bibliography{refs}



\end{document}